\documentclass{article}

\usepackage[preprint]{neurips_2022}
\usepackage{graphicx}

\usepackage[utf8]{inputenc} 
\usepackage[T1]{fontenc}    
\usepackage{hyperref}       
\usepackage{url}            
\usepackage{booktabs}       
\usepackage{amsfonts}       
\usepackage{nicefrac}       
\usepackage{microtype}      
\usepackage{xcolor}         

\title{Identifying Subgroups of ICU Patients Using End-to-End Multivariate Time-Series Clustering Algorithm Based on Real-World Vital Signs Data}

\author{%
    Tongyue Shi\textsuperscript{\normalfont 1,2}\quad
    Zhilong Zhang\textsuperscript{\normalfont 1}\quad
    Wentie Liu\textsuperscript{\normalfont 1}\quad 
    Junhua Fang\textsuperscript{\normalfont 2}\quad\\
    \textbf{Jianguo Hao}\textsuperscript{\normalfont 1}\quad
    \textbf{Shuai Jin}\textsuperscript{\normalfont 1}\quad
    \textbf{Huiying Zhao}\textsuperscript{\normalfont 3}\quad
    \textbf{Guilan Kong}\textsuperscript{\normalfont 1,4}\thanks{Guilan Kong is the corresponding author.} \quad
    \\ 
    \textsuperscript{1}National Institute of Health Data Science, Peking University, Beijing, China \\
    \textsuperscript{2}School of Computer Science and Technology, Soochow University, Suzhou, China \\
    \textsuperscript{3}Peking University People’s Hospital, Beijing, China\\
    \textsuperscript{4}Advanced Institute of Information Technology, Peking University, Hangzhou, China\\
    \texttt{tyshi@stu.pku.edu.cn, guilan.kong@hsc.pku.edu.cn} \\
}

\begin{document}

\maketitle

\begin{abstract}
This study employed the MIMIC-IV database as data source to investigate the use of dynamic, high-frequency, multivariate time-series vital signs data, including temperature, heart rate, mean blood pressure, respiratory rate, and SpO2, monitored  first 8 hours data in the ICU stay. Various clustering algorithms were compared, and an end-to-end multivariate time series clustering system called Time2Feat, combined with K-Means, was chosen as the most effective method to cluster patients in the ICU. In clustering analysis, data of  8,080 patients admitted between 2008 and 2016 was used for model development and 2,038 patients admitted between 2017 and 2019 for model validation. By analyzing the differences in clinical mortality prognosis among different categories, varying risks of ICU mortality and hospital mortality were found between different subgroups. Furthermore, the study visualized the trajectory of vital signs changes. The findings of this study provide valuable insights into the potential use of multivariate time-series clustering systems in patient management and monitoring in the ICU setting.
\end{abstract}

\section{Background}
The Intensive Care Unit (ICU) is a specialized medical facility that provides intensive monitoring and treatment for critically ill patients. ICU patients are characterized by severe illness and life-threatening conditions, requiring close monitoring and treatment. The changes in vital signs have multifaceted implications for patients. Existing research on patient subgroup analysis often focuses on single diseases and depends on cross-sectional analysis[1], and the value of dynamic multivariate time-series data of vital signs have not been utilized[2]. Therefore, there is a research gap in the literature about making full use of time-series vital sign data to explore subgroups in ICU patients for precision ICU care.

\section{Objectives}
This study aimed to use end-to-end multivariate time-series clustering algorithm to identify subgroups of ICU patients based on dynamic vital sign data recorded during the first 8 hours after ICU admission, and then to further explore the differences of prognoses in different patient subgroups. Overall, this study would make contributions to precision ICU care by classifying patients into different subgroups for more personalized clinical interventions.


\section{Methods}
In this study, the Medical Information Mart for Intensive Care (MIMIC)-IV database[3] was used as the data source. The dynamic, high-frequency vital sign data monitored in ICU during the first 8 hours was used for analysis. We used multivariate time-series clustering algorithms to cluster and group critically ill ICU patients first, and then analyzed the patient prognosis in different subgroups to help clinicians identify those patients with high mortality risk. All adult ICU patients in MIMIC-IV were included. In clustering analysis, data of patients admitted between 2008 and 2016 was used for model development and patients admitted between 2017 and 2019 for model validation. Variables including gender, age, race, height, weight, date of death (DoD), together with hourly monitored vital signs: temperature, heart rate, mean blood pressure, respiratory rate, and SpO2 were extracted. For patients having multiple ICU stays in one hospital admission, only the first ICU admission record was extracted. Patients were excluded if they had missing values for the extracted variables. Patient prognoses including ICU mortality, hospital mortality were analyzed for each patient subgroup. Finally, The elbow method and metrics including Davies-Bouldin Index (DBI) and Calinski-Harabaz Index (CHI) were used to determine the optimal number of clusters (\emph{k}) and optimal clustering algorithm[4].

\section{Results}

In this study, a total of 10,118 patients including 8,080 patients admitted between 2008 and 2016, and 2,038 patients admitted between 2017 and 2019 were included. Several clustering models, including Time2Feat[5] combined with k-Means, k-Shape, k-Medoids and Density-Based Spatial Clustering of Applications with Noise (DBSCAN) were employed for analysis, and finally the Time2Feat combined with k-Means model was selected as it had the best performance with CHI of 341.59 and DBI of 5.92. According to the elbow method, the optimal number of clusters is determined to be 3 (\emph{k}=3). In model development process, 8080 patients admitted from 2008 to 2016 were divided into three subgroups. In model validation process, 2038 patients admitted from 2017 to 2019 were divided into three subgroups as well. As depicted in Figure 1, vital sign trajectories in the three identified subgroups are similar in both the model development and validation datasets. And there are noticeable differences in the trajectories of heart rate, SpO2, temperature and respiratory rate among the three subgroups, while the average blood pressure trajectories show less apparent distinctions in the three subgroups. As to hospital mortality, on the dataset for model development, the risk ranked from highest to lowest were Subgroup2 (0.1092±0.005), Subgroup1 (0.0875±0.0104), and Subgroup3 (0.0867±0.0048); on the validation dataset, the risk showed consistent order: Subgroup2 (0.1245±0.0117), Subgroup1 (0.1218±0.0145), and Subgroup3 (0.1033±0.0113). Regarding the ICU mortality, the risk ranked from highest to lowest were Subgroup1 (0.0485±0.0079), Subgroup2 (0.0468±0.0034), and Subgroup3 (0.0242±0.0026) on the model development dataset, and the order was Subgroup2 (0.0436±0.007), Subgroup1 (0.0393±0.0086), and Subgroup3 (0.0234±0.0056) on the validation dataset. There was a slight ICU mortality difference between Subgroup1 and Subgroup2. Considering the smaller sample size in the validation dataset, there is a certain margin of error. However, both Subgroup1 and Subgroup2 had higher ICU mortality rates than the overall rate (0.0353±0.0041).

\begin{figure}[t]
\begin{center}
   \includegraphics[width=0.9\linewidth, height=1.3\linewidth]{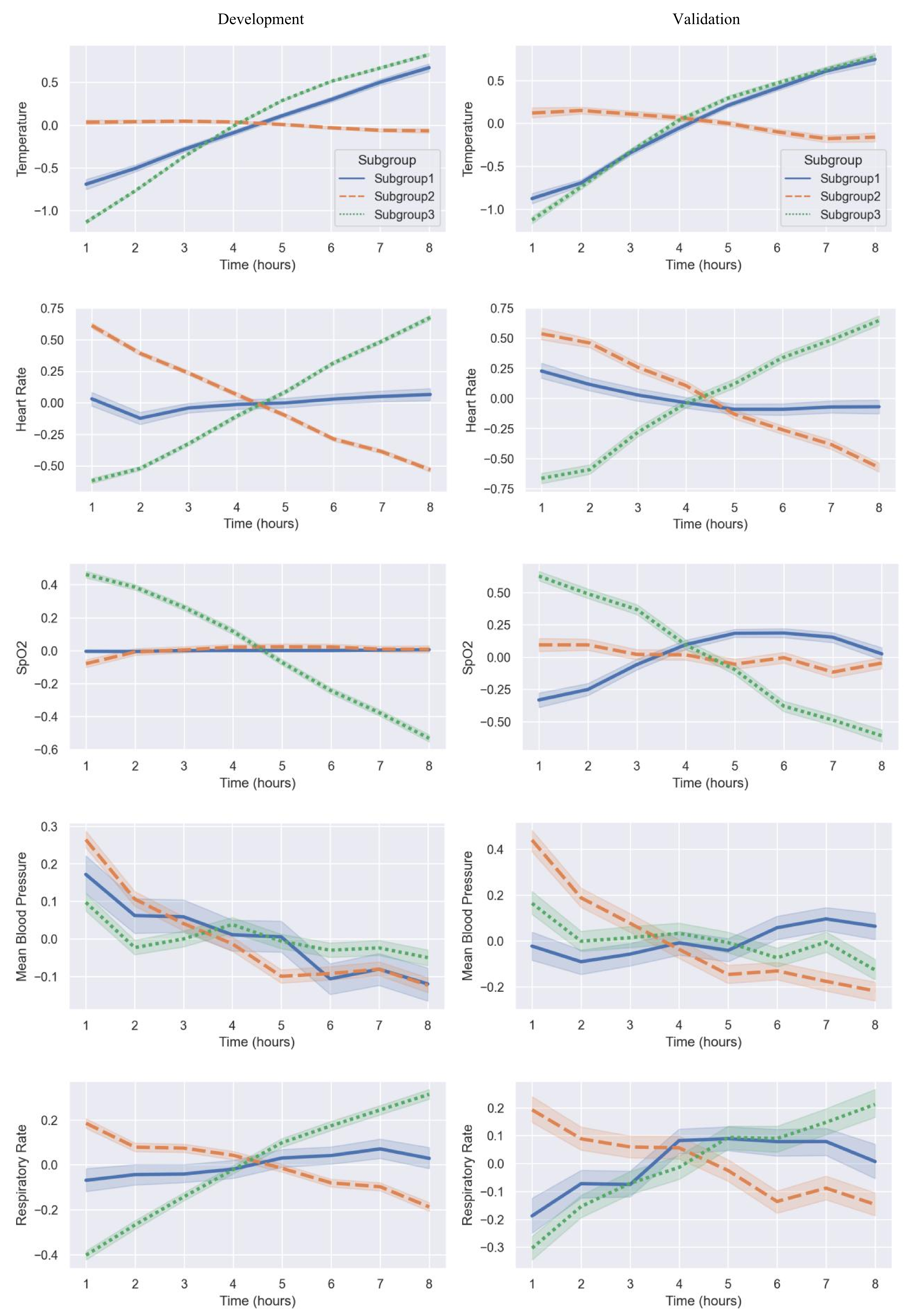}
\end{center}
   \caption{Trajectories of Vital Signs During the First 8 Hours in the Three Identified ICU Patient Subgroups.}
\vspace{-1em}
\label{fig1}
\end{figure}

\section{Conclusion}
The multivariate time-series data of vital signs monitored during the first 8 hours after ICU admission can reflect real conditions of patients and help to predict prognoses to some extent. By employing proper multi-variate time-series clustering algorithm to make second use of real-world vital sign data recorded in ICU could help clinicians to identify distinct patient subgroups with different mortality risks. The Time2Feat combined with k-Means method used in this study has shown satisfactory clustering performance. In the next step, we will generalize the Time-Series Clustering approach to other diseases and refine the model in practical applications.

\begin{ack}
This study was supported by Grants from the Zhejiang Provincial Natural Science Foundation of China (Grant No. LZ22F020014), National Key Research and Development Program of China (Grant No. 2018AAA0102100), Beijing Municipal Science \& Technology Commission (Grant No. 7212201), Humanities and Social Science Project of Chinese Ministry of Education (Grant No. 22YJA630036).
\end{ack}

\section*{References}
{
\small
[1]Liu K, Zhang X, Chen W, et al. Development and validation of a personalized model with transfer learning for acute kidney injury risk estimation using electronic health records[J]. JAMA Network Open, 2022, 5(7): e2219776-e2219776. doi:10.1001/jamanetworkopen.2022.19776.

[2]Tharakan S, Nomoto K, Miyashita S, et al. Body temperature correlates with mortality in COVID-19 patients[J]. Critical care, 2020, 24: 1-3. doi:10.1186/s13054-020-03045-8.

[3]Johnson A E W, Bulgarelli L, Shen L, et al. MIMIC-IV, a freely accessible electronic health record dataset[J]. Scientific data, 2023, 10(1): 1. doi:10.1038/s41597-022-01899-x.

[4]Kodinariya T M, Makwana P R. Review on determining number of Cluster in K-Means Clustering[J]. International Journal, 2013, 1(6): 90-95.

[5]Bonifati A, Buono F D, Guerra F, et al. Time2Feat: learning interpretable representations for multivariate time series clustering[J]. Proceedings of the VLDB Endowment, 2022, 16(2): 193-201. doi:10.14778/3565816.3565822.

}

\end{document}